%% file: main.tex
\documentclass[runningheads]{llncs}
\usepackage[T1]{fontenc}
\usepackage{graphicx}
\usepackage{booktabs}
\usepackage[misc]{ifsym}

\usepackage{graphicx}
\usepackage{algorithm}
\usepackage{algpseudocode}

\usepackage{bbding}
\usepackage{CJKutf8}
\usepackage{enumitem}
\usepackage{tabularray}
\usepackage{makecell}
\usepackage{subfigure}
\usepackage{latexsym}
\usepackage{xcolor}
\usepackage{caption}
\usepackage{threeparttable}
\usepackage{booktabs}
\usepackage{multirow}
\usepackage[normalem]{ulem}
\usepackage{hyperref}
\usepackage{float}
\input{math_commands.tex}

\usepackage{threeparttable}
\usepackage{multirow}
\usepackage{colortbl}
\definecolor{mygray}{gray}{0.85}

\newcommand{\model}{VIGraph\xspace}

\usepackage[misc]{ifsym}
\usepackage{mwe}

\newcommand{\vpara}[1]{\vspace{0.04in}\noindent\textbf{#1}\xspace}
\begin{document}

\title{VIGraph: Generative Self-supervised Learning for Class-Imbalanced Node Classification}
\titlerunning{VIGraph}
\institute{}
\author{Yulan Hu\inst{1,2} \and
  Sheng Ouyang\inst{1} \and
  Zhirui Yang\inst{1} \and
  Yong Liu\inst{1}
}

\authorrunning{Hu, Ouyang et al.}

\institute{Gaoling School of Artificial Intelligence, Renmin University of China, China \and
  Kuaishou Technology, China
  \email{\{huyulan,ouyangsheng,yangzhirui,liuyonggsai\}@ruc.edu.cn}}
%



\maketitle              

\begin{abstract}
Class imbalance in graph data presents significant challenges for node classification. While existing methods, such as SMOTE-based approaches, partially mitigate this issue, they still exhibit limitations in constructing imbalanced graphs. Generative self-supervised learning (SSL) methods, exemplified by graph autoencoders (GAEs), offer a promising solution by directly generating minority nodes from the data itself, yet their potential remains underexplored. In this paper, we delve into the shortcomings of SMOTE-based approaches in the construction of imbalanced graphs. Furthermore, we introduce VIGraph, a simple yet effective generative SSL approach that relies on the Variational GAE as the fundamental model. VIGraph strictly adheres to the concept of imbalance when constructing imbalanced graphs and innovatively leverages the variational inference (VI) ability of Variational GAE to generate nodes for minority classes. VIGraph introduces comprehensive training strategies, including cross-view contrastive learning at the decoding phase to capture semantic knowledge, adjacency matrix reconstruction to preserve graph structure, and alignment strategy to ensure stable training. VIGraph can generate high-quality nodes directly usable for classification, eliminating the need to integrate the generated nodes back to the graph as well as additional retraining found in SMOTE-based methods. We conduct extensive experiments, results from which demonstrate the superiority and generality of our approach. 
\end{abstract}

\section{Introduction}~\label{intro}
Node classification is a fundamental task in graph learning~\cite{arora2020survey}, which can be effectively addressed by various types of Graph Neural Networks (GNNs)~\cite{gcn}. However, real-world graph data often demonstrates class imbalance, wherein the number of nodes in one or more classes is significantly lower than in others, posing a challenge for GNNs in such scenarios.

\begin{figure}[t!]
  \centering
  \subfigure[]{
    \label{intro:moti:construct0}
    \includegraphics[width=0.28\textwidth]{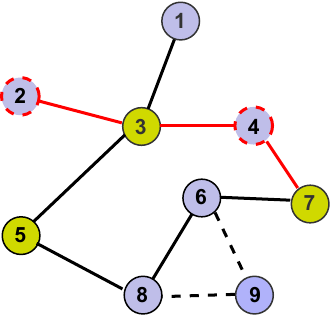}}
    \subfigure[]{
    \label{intro:moti:construct1}
    \includegraphics[width=0.28\textwidth]{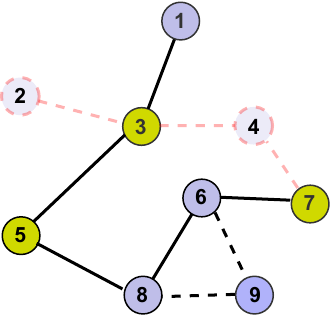}}    
  \subfigure[]{
    \label{intro:moti:accuracy}
    \includegraphics[width=0.36\textwidth]{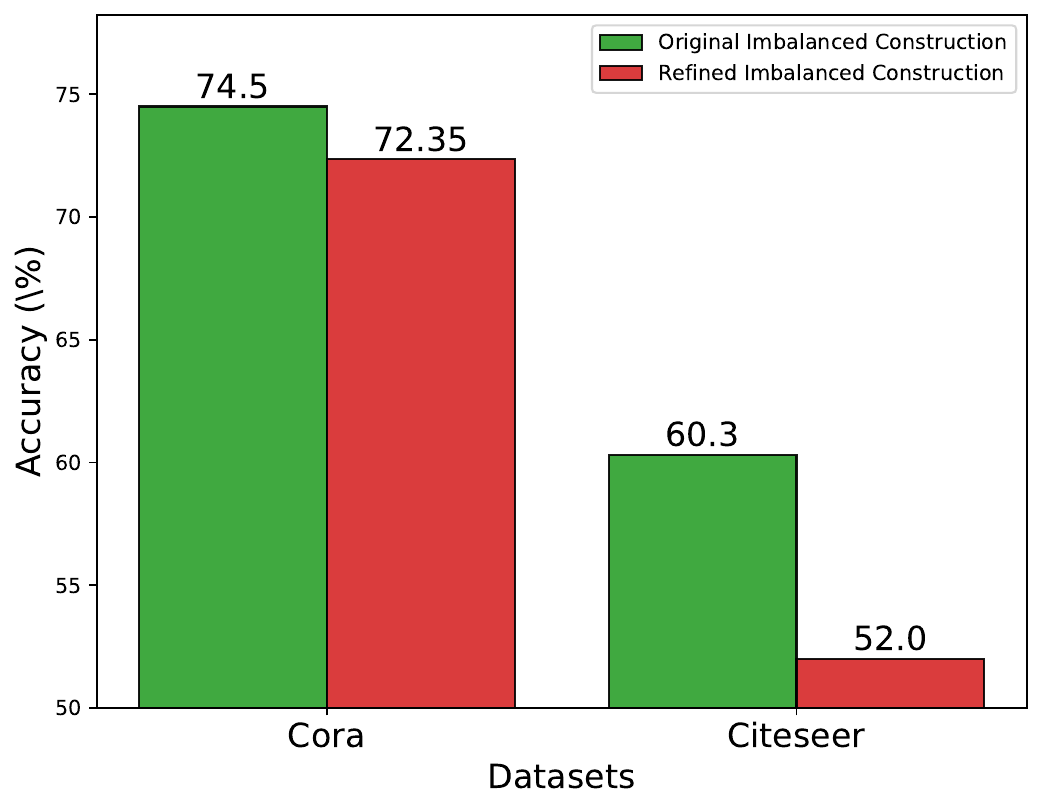}}
  \caption{The explanation of two kinds of imbalance graph construction. Fig~\ref{intro:moti:construct0} depicts the construction method adopted by the SMOTE-based approaches, wherein specific nodes (node 2 and node 4) are masked while the edges connected to them are retained. Conversely, Fig~\ref{intro:moti:construct1} illustrates the rigorous construction, where both the nodes and the edges connected to them are removed. Additionally, Fig~\ref{intro:moti:accuracy} showcases the change in accuracy of GraphSmote observed under these two construction methods on Cora and CiteSeer.
  }
  \label{intro:moti}
  \setlength{\abovecaptionskip}{0cm}
\end{figure}

In traditional machine learning, the problem of class imbalance has been extensively studied~\cite{classimbtrad_1,classimbtrad_0}. However, due to the non-Euclidean characteristics, directly applying these methods to graph data is non-trivial. To date, methods for addressing the class imbalance problem in graphs can be broadly categorized into two groups: re-sampling and re-weighting methods. Re-sampling methods tackle the issue by either upsampling nodes of the minority class or downsampling nodes of the majority class. SMOTE-based approaches~\cite{mixup} are commonly employed for the sampling process. Representative upsampling methods include GraphSmote~\cite{graphsmote}, GraphMixup~\cite{graphmixup}, and GraphENS~\cite{graphens}. These methods perform interpolation between nodes to synthesize additional minority nodes. In contrast, ImGCL~\cite{zeng2022imgcl} focuses on downsampling, gradually reducing the number of majority nodes during the training process. On the other hand, re-weighting methods address the problem by assigning large weights to minority classes or adjusting the inference logits for minority classes~\cite{li2019gradient,li2019dice,lin2017focal}. Additionally, ReNode~\cite{renode} focuses on detecting topological imbalances for individual nodes and then adjusts the weight of the identified nodes during the training process. The procedures of upsampling, especially the SMOTE-based methods, are rather intuitive and can be divided into three sub-steps: (1) Synthesizing nodes for minority classes based on the pre-trained graph embedding. (2) Integrating the synthesized nodes back to obtain a balanced graph. (3) Retraining the graph embedding based on the balanced graph.

However, this paradigm exhibits several limitations. First, there is a critical drawback with SMOTE-based methods~\cite{graphsmote,graphmixup,graphens} during the synthesis of minority nodes. Figure~\ref{intro:moti:construct0} illustrates the process of constructing imbalanced graphs using these approaches. Due to the lack of benchmark datasets for imbalanced graphs, existing SMOTE-based methods adopt the conventional practice of manually disrupting the node quantity distribution of balanced graphs to obtain imbalanced graphs. Specifically, these methods first obtain node representations with GNNs based on the original balanced graph. Then, they remove a proportion of nodes from predefined minority classes to create the imbalanced scenario. Afterward, they synthesize new nodes based on the node representations learned from the balanced graph. We believe this approach lacks rigor because the node representations trained from a balanced and an imbalanced graph are obviously different due to the different message passing process under distinct graph structures~\cite{balcilar2021breaking}. We argue that \textbf{the node representations used for upsampling should be trained on the imbalanced graph rather than on the original balanced graph} to better simulate realistic scenarios. We conducted experiments based on GraphSmote~\cite{graphsmote} to reveal the negative impact brought by the improperly constructed imbalanced graphs. As shown in Figure~\ref{intro:moti:construct1}, GraphSmote's performance has declined to varying degrees on both datasets under the rigorous setting, indicating that the incorrectly constructed imbalanced graphs wrongly boost the overall performance. Second, the SMOTE-based methods perform interpolation between existing minority nodes, which may synthesize outliers inconsistent with the existing distribution of minority nodes. Third, the integrating and retraining process is tedious and requires relatively complex design. The synthesized nodes need to be integrated back into the imbalanced graph, which involves measuring the distances of each synthesized node from its neighbors. Additionally, the balanced graph needs to be retrained for refined node representation.

The above analysis reveals several limitations of SMOTE-based methods in synthesizing minority nodes. In recent years, generative self-supervised learning (SSL) models have achieved success in diverse fields~\cite{mae,clip,simcse}, yet their potential in addressing class-imbalanced node classification remains unexplored. Inspired by this, we aim to generate, rather than synthesize, the minority nodes to tackle the problem. To achieve this, three challenges need to be addressed. \textbf{First}, how to construct an imbalanced graph that strictly adheres to the imbalance setting? In contrast to SMOTE-based methods, we disrupt a proportion of nodes to obtain an imbalanced graph from the outset. Furthermore, we not only remove nodes but also eliminate the edges connected to the discarded nodes to sever the message passing path. This approach ensures that the node representations are learned from the imbalanced graph, maximizing the simulation of real-world imbalanced graphs and avoiding the limitations of SMOTE-based methods. \textbf{Second}, is there a model or structure that can serve as the foundation? The variational graph autoencoder (VGAE)~\cite{kipf2016variational} represents an early SSL attempt to tackle the link prediction task. Built upon the variational autoencoder (VAE), VGAE applies variational inference (VI) to graph learning, which can also serve as a tool to generate new nodes. We employ VGAE as the backbone model and leverage VI to generate minority nodes, ensuring that the generated nodes are learned from the existing nodes. \textbf{Third}, how to improve the quality of the generated nodes? To generate nodes that seamlessly combine with the existing nodes, we design comprehensive strategies to refine both the graph features and structure, including a structure reconstruction strategy and a novel siamese contrastive learning strategy at the decoding phase. Additionally, we leverage the alignment strategy with Kullback-Leibler (KL) to ensure stable training. These strategies ultimately compel the encoder to enhance its encoding ability, resulting in higher-quality node representations.

After successfully addressing the above three challenges, we propose VIGraph, the first generative SSL model to tackle class-imbalanced node classification. To summarize, our contributions include:

\begin{itemize}     
    \item We reexamine the challenges of class-imbalanced node classification, identifying the critical limitations of SMOTE-based methods. 
    \item We introduce VIGraph, a novel generative SSL model that leverages generative SSL to generate minority nodes. To the best of our knowledge, this is the first attempt to apply generative SSL to class-imbalanced graph node classification.
    \item We conduct extensive experiments on multiple real-world datasets, and the results validate the effectiveness of VIGraph.
\end{itemize}

\section{Related Work}

\subsection{Class Imbalance Node Classification}
The approaches of addressing this problem can be roughly divided into two categories~\cite{ma2023class}: Re-sampling and Re-weighting.

\vpara{Re-sampling approaches.} Re-sampling approaches aim to balance the distribution of labeled data to tackle the imbalance problem. GraphSMOTE~\cite{graphsmote} utilizes the SMOTE algorithm to synthesize nodes for the minority classes, with a graph structure reconstruction model trained to predict the edges of the newly synthesized nodes. However, determining the upsampling ratio for GraphSMOTE can be challenging. To address this issue, GraphMixup~\cite{graphmixup} devises a reinforcement mixup mechanism to adaptively determine the upsampling ratio. GraphENS~\cite{graphens} synthesizes a new node for the minority classes and its ego graph via the ego graph of a minority class node and an arbitrary node, based on neighbor sampling and saliency-based node mixing. In contrast to the above three methods, which adopt upsampling techniques, ImGCL~\cite{zeng2022imgcl} points out that graph contrastive learning methods are susceptible to imbalanced node classification settings and employs the node centrality-based progressively balanced sampling method to balance the distribution of labeled data by downsampling the majority nodes. ImGAGN~\cite{qu2021imgagn} and SORAG~\cite{duan2023anonymity} adopt an adversarial generation method to generate nodes as well as their topology for the minority classes, representing a different type of upsampling method. In addition, SPARC~\cite{zhou2018sparc} and SETGNN~\cite{juan2021exploring} expand the set of nodes for the minority classes by assigning pseudo-labels to unlabeled nodes with self-training methods.

\vpara{Re-weighting approaches.} Re-weighting approaches aim to refine algorithms to address the imbalance problem. ACS-GNN~\cite{ma2022attention} assigns individual weights to majority class and minority class samples via an attention mechanism and cost-sensitive techniques. ReNode~\cite{renode} introduces a conflict detection-based topology relative location metric (Totoro) to measure the degree of topological imbalance for each node, based on which the weight of each node in the loss function is adjusted. TOPAUC~\cite{chen2022unified} is another reweighting method that utilizes a topology-aware importance learning (TAIL) mechanism for AUC optimization. TAM~\cite{pmlr-v162-song22a} designs anomalous connectivity-aware margin and anomalous distribution-aware margin to adjust the logits of the model output.

\subsection{Generative Self-Supervised Graph Learning}
According to the reconstruction paradigm, generative SSL methods can be classified into two categories: graph autoencoder (GAE) methods and graph autoregressive methods~\cite{wu2021self}.

\vpara{Graph autoencoder.} GAE generally comprise an encoder to map the input graph into hidden representation, followed by another decoder to reconstruct the input graph from the hidden representation. GAE and VGAE~\cite{kipf2016variational} represent pioneering generative SSL methods aiming to reconstruct the graph structure with a simple inner product decoder. ARGA and ARVGA~\cite{pan2019adversarially} use GAE as the backbone structure, additionally forcing the latent representation to match a prior distribution through an adversarial training scheme. GALA~\cite{Park_2019_ICCV} proposes a symmetric graph convolutional autoencoder architecture, which includes a Laplacian smoothing encoder and a Laplacian sharpening decoder as counterparts of Laplacian smoothing. GraphMAE~\cite{graphmae} introduces the masked autoencoder (MAE) to the field of graph learning, focusing on reconstructing node features using masked graph autoencoder. GraphMAE also develops several innovative training strategies, including masked feature reconstruction, scaled cosine error (SCE), and re-mask decoding. HGMAE~\cite{hgmae} introduces the masked graph autoencoder into the field of heterogeneous graph learning, achieving promising performance with comprehensive training strategies.

\vpara{Graph autoregressive.} Graph autoregressive methods undertake the reconstruction process in a sequential manner. GraphRNN~\cite{you2018graphrnn} interprets graphs varying in node order as specific sequences, thus developing an autoregressive generative model based on these sequences. GCPN~\cite{you2018graph} harnesses the capabilities of reinforcement learning to generate molecular graphs autoregressively. GPT-GNN~\cite{hu2020gpt} progressively generates node attributes and relationships through two central components: attribute generation and edge generation. Additionally, CCGG~\cite{ommi2022ccgg} constitutes a class-conditional autoregressive graph generation model wherein class-specific data is woven into the generation of the graph structure.

\section{Methodology}
\begin{figure*}[h]
  \centering
  \includegraphics[width=1.0\textwidth]{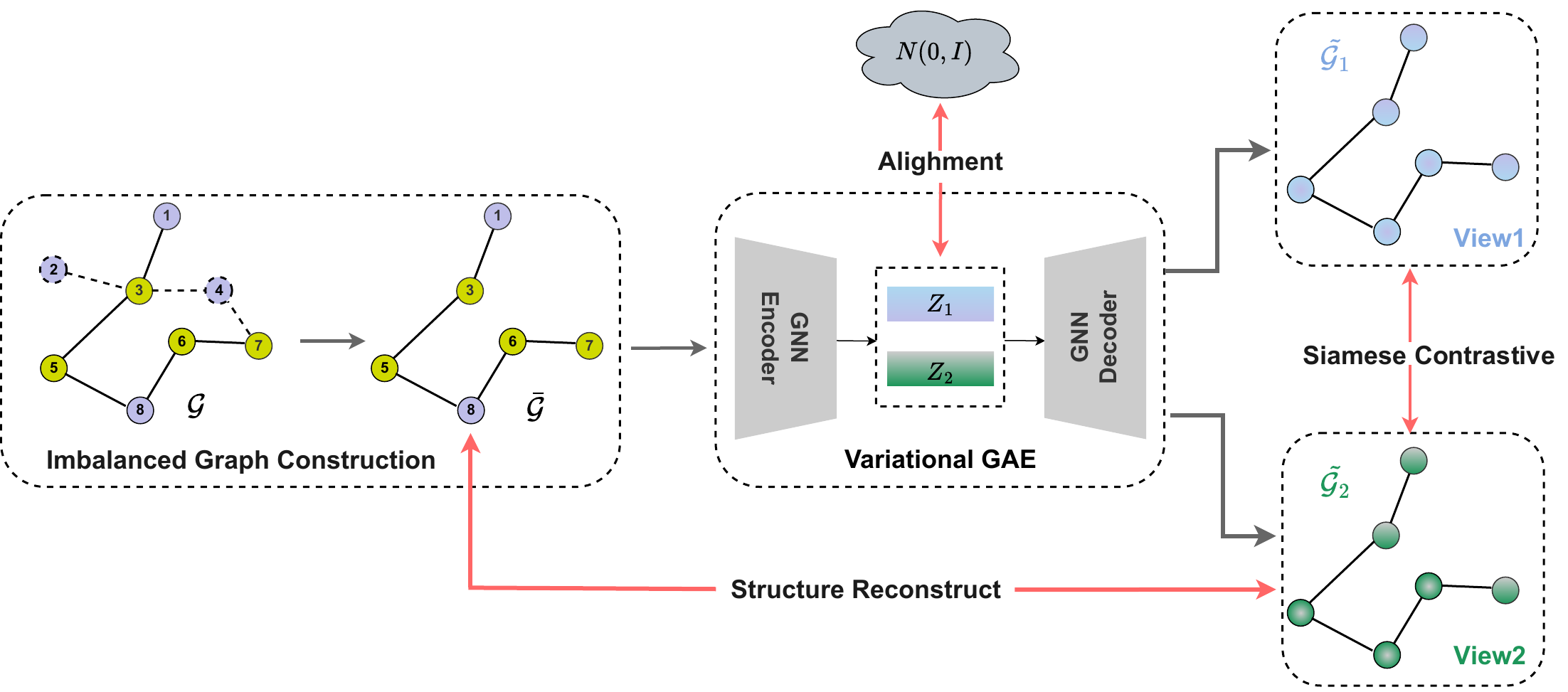}
  \caption{
  The Overview of VIGraph. Initially, the input graph $\mathcal{G}$ is balanced. We manually remove a portion of nodes along with their linked edges to construct an imbalanced graph, denoted as $\bar{\mathcal{G}}$. Subsequently, the variational GNN encoder processes the imbalanced graph as input, performing pairwise encoding to obtain two latent representations, $Z_1$ and $Z_2$. Following this, the GNN decoder reconstructs the imbalanced graph based on $Z_1$ and $Z_2$, resulting in $\tilde{\mathcal{G}}_1$ and $\tilde{\mathcal{G}}_2$, respectively. Additionally, we introduce three strategies to enhance training, which include structure reconstruction between $\bar{\mathcal{G}}$ and $\tilde{\mathcal{G}}_2$ (or $\tilde{\mathcal{G}}_1$), cross-view contrastive learning, and distribution alignment between the latent representation and the posterior distribution.}
  \label{fig:overview}
\end{figure*}

We formally present VIGraph in this section. The overall architecture of VIGraph is depicted in Figure~\ref{fig:overview}. Following the pre-training process, we leverage the ability of variational inference to generate new nodes for minority classes.

\subsection{Preliminaries}\label{model:pre}
\vpara{Notations.}
Given a graph \(\mathcal{G} = (\mathcal{V}, \mathcal{E}, \mathcal{X}, \mathcal{Y}, \mathcal{A})\), where \(\mathcal{V}\) is the set of \(N\) nodes and \(\mathcal{E} \subseteq \mathcal{V} \times \mathcal{V}\) is the set of edges. \(\mathcal{X} \in \mathbb{R}^{N \times F_s}\), where \(F_s\) is the dimension of the input node feature. Each node \(v \in \mathcal{V}\) is associated with a feature vector \(x_v \in \mathcal{X}\), and each edge \(e_{u,v} \in \mathcal{E}\) denotes a connection between node \(u\) and node \(v\). \(\mathcal{Y} \in \mathbb{R}^n\) represents the class information for the nodes with label information in \(\mathcal{G}\). The graph structure can also be represented by an adjacency matrix \(\mathcal{A} \in \{0,1\}^{N \times N}\) with \(A_{u,v} = 1\) if \(e_{u,v} \in \mathcal{E}\) and \(A_{u,v} = 0\) if \(e_{u,v} \notin \mathcal{E}\). 

\vpara{Problem statement.}
Suppose there are \(k\) node classes \(\mathcal{C} = (\mathcal{C}_1,...,\mathcal{C}_k)\) in graph \(\mathcal{G}\), where \(|\mathcal{C}_i|\) represents the number of nodes in class \(i\). The imbalance ratio, \(\lambda = \frac{\min_i |\mathcal{C}_i|} {\max_i |\mathcal{C}_i|}\), measures the degree of class imbalance. In the semi-supervised setting, labeled nodes are far fewer than unlabeled ones. We denote the labeled set as \(\mathcal{D}_L = (\mathcal{V}_L, \mathcal{X}_L, \mathcal{Y}_L)\) and the unlabeled set as \(\mathcal{D}_U = (\mathcal{V}_U, \mathcal{X}_U, \mathcal{Y}_U)\), with imbalance typically observed in \(\mathcal{D}_L\). Given the imbalanced graph \(\mathcal{G}\) and a labeled node subset, VIGraph aims to generate high-quality nodes for minority classes, denoted as \(\mathcal{D}_G = (\mathcal{V}_G, \mathcal{X}_G, \mathcal{Y}_G)\). After augmenting \(\mathcal{D}_L\) with \(\mathcal{D}_G\), we obtain the synthesized set \(\mathcal{D}_S = \mathcal{D}_L \cup \mathcal{D}_G\), where \(\mathcal{D}_S\) exhibits quantity balance across different classes.

\subsection{Imbalanced Graph Construction}\label{graph:construct}
As discussed in Section~\ref{intro}, the previous SMOTE-based up-sampling approaches for imbalanced node classification can be summarized into three steps.

\vpara{Synthesizing.} First, the original balanced graph \(\mathcal{G}\) is encoded by GNNs, obtaining representation \(\mathcal{H}\). Subsequently, a subset of classes is selected as minority classes (denoted as \(C_m\)), with a portion of nodes in \(C_m\) designated as minority nodes, resulting in an imbalanced graph \(\bar{\mathcal{G}}\). Mixup interpolation is then performed between minority nodes based on \(\mathcal{H}\), yielding the synthesized minority nodes \(\mathcal{V}_G\). \vpara{Integrating.} The distances between the synthesized nodes \(\mathcal{V}_G\) and the nodes in \(\bar{\mathcal{G}}\) are calculated, and connectivity between \(\mathcal{V}_G\) and \(\bar{\mathcal{G}}\) is determined based on a predefined threshold. Specifically, an edge is created if the distance is smaller than the threshold; otherwise not. Following this integration process, a balanced graph is obtained, denoted as \(\tilde{\mathcal{G}}\). \vpara{Retraining.} The manually constructed balanced graph, \(\tilde{\mathcal{G}}\), undergoes encoding by GNNs to obtain representation \(\tilde{\mathcal{H}}\), which is subsequently utilized for node classification.

We believe that the construction process, particularly the generation of nodes for minority classes, is problematic. The representation for node generation should be based on the imbalanced graph \(\bar{\mathcal{G}}\) rather than the original balanced graph \(\mathcal{G}\). Additionally, the discarded nodes along with the edges connected to them should be removed to prevent their participation in message passing. Unfortunately, the SMOTE-based methods neglect this critical requirement and erroneously utilize \(\mathcal{H}\) for feature interpolation, leading to biased class-imbalanced node classification results.

In contrast to previous methods, we introduce a more rigorous approach to constructing the imbalanced graph. Primarily, we initiate the construction process by directly forming the imbalanced graph, upon which all subsequent strategies are developed. Furthermore, we not only remove nodes but also discard connecting edges to mitigate the impact of message passing. Consequently, the resulting imbalanced graph represents a subset of the original graph, adhering closely to realistic imbalanced scenarios. For simplicity, we will not explicitly distinguish between the original graph and the constructed subgraph in the following sections. By default, we will refer to the subgraph as the graph.

\subsection{Variational Graph Autoencoder}~\label{model:encoder}
\vpara{Stochastic properties encoding.}
VIGraph employs a single-layer GNN to encode the imbalanced graph as follows:
\begin{equation}
  \mathbf{H} = \text{ReLU}(\text{GNN}(\mathbf{X})),
  \label{encode}
\end{equation}
Here, $\text{ReLU}(\cdot)=\max (0, \cdot)$, and $\mathbf{H} \in \mathbb{R}^{N \times F_d}$. $F_d$ denotes the hidden feature dimension. Subsequently, VIGraph focuses on learning the posterior distribution $p_{\theta}(\mathbf{Z}|\mathbf{H},\mathbf{A})$ for generating latent variable $\mathbf{Z}$, where $\mathbf{z}_i$ represents the latent variable of node $i$. To achieve this, we introduce two auxiliary GNN layers, $\text{GNN}_{\boldsymbol{\mu}}$ and $\text{GNN}_{\boldsymbol{\sigma}}$, to independently learn the mean and logarithm of variance for each node, denoted as $\boldsymbol{\mu}$ and $\log {\boldsymbol{\sigma}}$. Due to the intractability of computing the true posterior, we introduce a new approximation function $q_{\phi}$, parameterized by $\phi$, to approximate the posterior given input $\mathbf{H}$ and $\mathbf{A}$~\cite{vgae}. Following VGAE~\cite{vgae}, we adopt the variational distribution in the form:
\begin{equation}
  q_{\phi}(\mathbf{Z} \mid \mathbf{H}, \mathbf{A}) = \prod_{i=1}^N q_{\phi}\left(\mathbf{z}_i \mid \mathbf{H},
   \mathbf{A}\right) = \prod_{i=1}^N \mathcal{N}\left(\mathbf{z}_i \mid 
   \boldsymbol{\mu}_i, \text{diag}(\boldsymbol{\sigma}_i^2)\right).
\end{equation}

\vpara{Pairwise reparameterization.} Once the stochastic variables of the latent representation are obtained, we can utilize the reparameterization technique~\cite{vgae} to sample from these latent variables. Specifically, sampling from the variational posterior $\mathbf{z}_i\sim q_{\phi}(\mathbf{z}_i|\mathbf{H}, \mathbf{A})$, is equivalent to sampling from a standard normal distribution:

\begin{equation}
  \mathbf{z}_i = \boldsymbol{\mu}_i + \boldsymbol{\sigma}_i \odot \boldsymbol{\epsilon}; \quad \boldsymbol{\epsilon} \sim \mathcal{N}(0,\mathbf{I}),
  \label{repam}
\end{equation}
where $\odot$ denotes element-wise multiplication, and $\epsilon$ is sampled from the standard normal distribution.

Unlike common practice, we employ pairwise reparameterization, yielding two latent variables for subsequent cross-view contrastive learning during the decoding phase, which we will elaborate on in Section~\ref{model:decoder}. Specifically, this strategy entails randomly sampling two variables from a standard normal distribution, then generating new representations following Equation~\ref{repam}, resulting in the latent representations $\mathbf{Z}_1$ and $\mathbf{Z}_2$. Indeed, given the stochastic variable of the target nodes, VIGraph can generate an arbitrary number of new latent representations. This concept serves as the foundation of our approach to generating samples for minority classes.

To ensure the proximity between the real posterior distribution and the approximate posterior distribution, we utilize the Kullback-Leibler (KL) divergence as the constraint between them. The Evidence Lower Bound (ELBO) is defined as follows:
\[
\mathcal{L}_{\text{ELBO}} = \mathbb{E}_{\mathbf{Z} \sim q_{\phi}(\mathbf{Z} \mid \mathcal{H}, \mathcal{A})} [\log p(\mathcal{H} \mid \mathbf{Z})] - \text{KL}(q_{\phi}(\mathbf{Z} \mid \mathcal{H}, \mathcal{A}) \| p_{\theta}(\mathbf{Z})).
\]

This formulation ensures that the model learns a latent representation $\mathbf{Z}$ that can effectively reconstruct the input data $\mathcal{H}$ while minimizing the discrepancy between the approximate posterior distribution $q_{\phi}(\mathbf{Z} \mid \mathcal{H}, \mathcal{A})$ and the prior distribution $p_{\theta}(\mathbf{Z})$.

\subsection{Comprehensive Decoding Strategies}~\label{model:decoder}
The GAEs typically utilize a reconstruction criterion to encapsulate the intrinsic characteristics of the input graph, with the decoder's configuration adapting based on downstream tasks~\cite{vgae,graphmae,hgmae}. Nevertheless, graph data differs from image data in that it comprises both feature and structural characteristics simultaneously. To fully leverage the underlying semantic information while retaining structural details, we have devised comprehensive training strategies aimed at capturing both semantic and structural knowledge within the graph.

\vpara{Structure reconstruction.} We aim to reconstruct the adjacency matrix from the decoded output and compare it with $\mathcal{A}$. To accomplish this, we employ a simple inner dot product as the structure decoder, with the reconstruction procedure formulated as follows:
\begin{equation}
  \hat{A}_{v, u} = \rho{\left({\hat{x}}_v \cdot \hat{x}_u^T\right)},
  \label{recon_eq}
\end{equation}
where $\hat{X}$ denotes the decoded features, $\hat{A}$ signifies the reconstructed adjacency matrix, $\hat{A}_{v, u}$ indicates the connectivity between nodes $u$ and $v$, and $\rho$ represents the sigmoid activation function.

The comparison between $\mathcal{A}$ and $\hat{A}$ can be regarded as a binary classification problem, where $\mathcal{A}$ serves as the ground truth while $\hat{A}$ represents the predicted results. Thus, we employ the cross-entropy ($\operatorname{CE}$) criterion to quantify the difference between $\mathcal{A}$ and $\hat{A}$. The inter-node connectivity effectively acts as an indicator of the class, with 1 indicating the presence of connectivity between two nodes and 0 indicating none. Due to sparsity, the number of negative samples (class 0) in the graph significantly outweighs the number of positive samples. To address this imbalance, we adopt a weighted version of the $\operatorname{CE}$ loss. For each row $r$ in the adjacency matrix $\mathcal{A}$, the weight assigned to the positive samples is computed as $W_{i} = (N - P_{i}) / P_{i}$, where $P_{i}$ represents the number of positive samples in row $r$. The final reconstruction loss is computed as follows:
\begin{equation}
  \mathcal{L}_{\text{rec}} =-\frac{1}{N^2} \sum_{i=1}^{N} \sum_{j=1}^{N} \left 
  [ W_i\left (\mathcal{A}_{i,j} \ln \hat{A}_{i,j} + (1 - \mathcal{A}_{i,j}) \ln (1 - \hat{A}_{i,j}) \right )\right ].
  \label{rec:loss}
\end{equation}

\vpara{Siamese contrastive decoding.} For GAE, as the semantic decoding quality of the objective gradually improves, it forces the encoder to generate more accurate graph representations, thereby enhancing overall GAE learning. Graph contrastive learning (GCL), as another form of SSL, aims to improve congruence at the node level between separate representations. At the decoding phase, we introduce GCL to capture cross-view semantic knowledge. Here, we employ a simple single-layer $\operatorname{MLP}$ as the feature decoder. Based on the two latent representations, $\mathbf{Z}_1$ and $\mathbf{Z}_2$, we use $\operatorname{MLP}$ to map them into two decoded representations, denoted by $\tilde{\mathbf{X}}_1$ and $\tilde{\mathbf{X}}_2$, respectively:

\begin{equation}
  \tilde{\mathbf{X}}_1 = \operatorname{ReLU}(\operatorname{MLP}(\mathbf{Z}_1)), 
  \quad
  \tilde{\mathbf{X}}_2 = \operatorname{ReLU}(\operatorname{MLP}(\mathbf{Z}_2)).
  \label{contr:7}
\end{equation}

We treat the two decoded representations as Siamese views, enabling node-level cross-view contrastive learning~\cite{grace}. Specifically, we consider both inter-view and intra-view contrastiveness. For a single node $x$, its representation in the first view, denoted as $\tilde{\mathbf{x}}_1$, serves as the anchor, while the representation of the same node in the second view, denoted as $\tilde{\mathbf{x}}_2$, represents the positive sample. Other nodes' representations in both views are considered negative samples. Thus, the positive pair can be denoted as $(\tilde{\mathbf{x}}_1, \tilde{\mathbf{x}}_2)^+$, and the negative pairs as $(\tilde{\mathbf{x}}_1, \tilde{\mathbf{v}}_1)^- \cup (\tilde{\mathbf{x}}_1, \tilde{\mathbf{v}}_2)^-$, where $\tilde{\mathbf{v}}$ indicates the representation of nodes in the Siamese views. 

To quantify the relationship between these representations, we introduce a cosine similarity function $\eta(\tilde{\mathbf{x}}_1, \tilde{\mathbf{x}}_2)$ to calculate the similarity score between nodes. Formally, we define the pairwise objective for each positive pair $(\tilde{\mathbf{x}}_1, \tilde{\mathbf{x}}_2)$ as follows:
\begin{small}
\begin{equation}
    \begin{aligned}
        & \mathcal{L}_{gcl}\left(\tilde{\mathbf{x}}_1, \tilde{\mathbf{x}}_2\right)                                                                                                                                   \\
      = & \log \frac{e^{\eta\left(\tilde{\mathbf{x}}_1, \tilde{\mathbf{x}}_2\right) / \tau}}{e^{\eta\left(\tilde{\mathbf{x}}_1, \tilde{\mathbf{x}}_2\right) / \tau} +\sum_{i = k}   e^{\eta\left(\tilde{\mathbf{x}}_i, \tilde{\mathbf{v}}_k\right) / \tau} +\sum_{i \neq k} e^{\eta\left(\tilde{\mathbf{x}}_i, \tilde{\mathbf{v}}_k\right)/ \tau}},
      \label{l:gcl}
    \end{aligned}
\end{equation}
\end{small}
where $\tau$ indicates the temperature value. In this way, we develop the cross-view contrastive mechanism to strengthen the semantic information.

\subsection{Minority Node Generation}~\label{model:generate}
The training process outlined above facilitates the learning and extraction of node representations and characteristics across the entire graph. For minority classes, our approach initiates an iterative reparameterization based on the learned stochastic latent variables $\boldsymbol{\mu}$ and $\boldsymbol{\sigma}$ according to Equation~\ref{repam}, continuing until the number of samples in each class reaches equilibrium.VIGraph generates minority nodes in a absolutely distinct manner compared to previous methods, which in the first time make utilization of variational inference to \textbf{generate} rather than \textbf{synthesizing} new nodes.

The quality of the samples generated by VIGraph is ensured through two aspects. Firstly, the generation process relies on the stochastic characteristics of existing minority samples. This ensures that the distribution of newly generated samples aligns with that of the existing minority samples, mitigating the potential introduction of outliers that may arise from convex combinations of node features. Secondly, the meticulously designed training strategy ensures a thorough exploration of the graph's features, enabling the newly generated nodes to closely resemble the existing minority nodes. In contrast to prior methods~\cite{graphsmote,graphmixup,graphens} that involve linking the nodes in $\mathcal{D}_G$ back to the graph for further training, we directly integrate the generated nodes with the original labeled node set $\mathcal{D}_L$. This results in a unified set denoted as $\mathcal{D}_S$, which is concise and free from additional training steps.

\subsection{Training and Optimization}~\label{model:train}
Altogether, we jointly optimize both the ELBO, the structure reconstruction criterion and the Siamese contrastive loss, the final optimization function is:
\begin{equation}
  \mathcal{L} = \alpha\mathcal{L}_{elbo}+ \beta\mathcal{L}_{rec} + \gamma\mathcal{L}_{gcl},
\end{equation}
where $\alpha$, $\beta$ and $\gamma$ are hyper-parameters to mediate each loss term.

\section{EXPERIMENTS}
In this section, we conduct extensive experiments to answer the following three research questions (RQ): \textbf{RQ1}: How does the VIGraph perform against other SOTA methods? \textbf{RQ2}: Does VIGraph exhibit robustness across varying class-imbalance ratios?  \textbf{RQ3}: What is the impact of different components and parameters on the overall performance of VIGraph?
\subsection{Experimental Setups}
To comprehensively evaluate the effectiveness of VIGraph, we conduct experiments on five real-world datasets, comprising three citation network datasets: Cora, CiteSeer, and PubMed~\cite{gcn}, a co-purchase graph, AmazonPhoto~\cite{photo}, and the Wikipedia-based graph Wiki-CS~\cite{huang2017label}, the full data information is introduced in Appendix. Following the methodology outlined in~\cite{graphsmote,renode}, we designate half of the available classes as minority classes for the three citation networks. Subsequently, we adjust the number of nodes in the minority classes according to the imbalance ratio, $\lambda$. Assuming the dataset consists of $|\mathcal{C}|$ labeled classes, each containing $N_c$ nodes, we select $\lfloor \frac{|\mathcal{C}|}{2} \rfloor$ classes as minority classes, with $\lambda \times N_c$ nodes assigned to each minority class. For AmazonPhoto, which comprises eight classes with varying numbers of samples, we select four of the eight classes with relatively fewer samples and manually set the number of nodes in these four classes to 10\% of the number of nodes in the classes with the most samples. Regarding Wiki-CS, we follow the approach outlined in GraphMixup~\cite{graphmixup} to identify classes with fewer samples than the average and designate them as minority classes.
 
\vpara{Baselines.} To assess the universality and stability of VIGraph, we employ GCN~\cite{gcn} as the backbone models.  We compare VIGraph against eight SOTA methods grouped as follows:
\begin{itemize}
  \item Loss Criterion. Cross Entropy (\textbf{CE}): CE is utilized as the loss criterion for the backbone model.
  \item Re-weighting Methods. (1) Re-Weight (\textbf{RW})~\cite{huang2016learning}: This method adjusts the loss functions. (2) Focal loss (\textbf{FC})~\cite{lin2017focal}: FC focuses on difficult-to-classify examples. (3) Class Balanced (\textbf{CB})~\cite{cui2019class}: CB handles class imbalance by assigning different weights to different classes. (4) ReNode (\textbf{RN})~\cite{renode}: RN measures the degree of topological imbalance in nodes.
  \item Re-sampling Methods. (1) GraphSMOTE (\textbf{GS})~\cite{graphsmote}: GS synthesizes new samples for minority classes. (2) GraphMixup (\textbf{GM})~\cite{graphmixup}: GM utilizes mixup interpolation to generate new samples. (3) GraphENS (\textbf{GE})~\cite{graphens}: GE synthesizes new nodes and their ego graphs based on neighbor sampling.
\end{itemize}
For VIGraph, we introduce a variant denoted as $\text{VIGraph}^\spadesuit$. $\text{VIGraph}^\spadesuit$ constructs the imbalanced graph following previous methods such as GraphSMOTE~\cite{graphsmote} and GraphMixup~\cite{graphmixup}, which we consider less rigorous.

\vpara{Setting.} We utilize accuracy (\textbf{Acc}), balanced accuracy (\textbf{bAcc}), and macro F-score (\textbf{F1}) as the evaluation metrics. We set the hidden dimension to 128, while the output dimension is fixed at 64. For optimization, we employ the Adam optimizer~\cite{KingmaB14} with a learning rate ranging from 0.0005 to 0.01. All experiments are executed on a single NVIDIA V100 32GB GPU. The reported results are averaged over ten runs to compute the mean and standard deviation.
 
\vpara{Main results.} We present the evaluation results in Table~\ref{exp:main_res_1} and Table~\ref{exp:main_res_2} to address \textbf{RQ1}. The performance of all compared baselines is listed if reported in the original paper; otherwise, we reproduce the results based on provided code. Due to space constraints, we abbreviate our method as VG. Beside, we only display the results of \model using GCN~\cite{gcn} as the backbone model. Results using GraphSAGE~\cite{graphsage} are also provided in the appendix.
\begin{table}[htb]
\caption{The experiment results on Cora, CiteSeer and PubMed. The best results are underlined, the row of $\text{VG}^\spadesuit$ presents the results that not strictly adhering to the imbalanced graph construction, which are shown in italics.}
\begin{threeparttable}
\begin{tabular}{@{}cccc|ccc|cc@{}}
\toprule
\multirow{2}{*}{} & \multicolumn{3}{c|}{Cora}                                                & \multicolumn{3}{c|}{CiteSeer}                                            & \multicolumn{2}{c}{PubMed}               \\ \cmidrule(l){2-9} 
                         & \multicolumn{1}{c|}{Acc}      & \multicolumn{1}{c|}{bACC}       & F1      & \multicolumn{1}{c|}{Acc}      & \multicolumn{1}{c|}{bAcc}     & F1       & \multicolumn{1}{c|}{Acc}      & bAcc     \\ \midrule
CE                       & \multicolumn{1}{c|}{63.4±1.1} & \multicolumn{1}{c|}{60.2±1.0} & 51.3±3.6 & \multicolumn{1}{c|}{53.7±1.4} & \multicolumn{1}{c|}{52.8±0.8} & 51.0±1.8 & \multicolumn{1}{c|}{75.3±0.4} & 73.0±0.8 \\
RW                       & \multicolumn{1}{c|}{75.0±1.1} & \multicolumn{1}{c|}{75.8±1.0} & 73.8±0.8 & \multicolumn{1}{c|}{61.4±1.9} & \multicolumn{1}{c|}{58.6±1.3} & 57.3±1.8 & \multicolumn{1}{c|}{74.6±1.0} & 74.1±0.8 \\
FC                       & \multicolumn{1}{c|}{65.5±1.9} & \multicolumn{1}{c|}{61.7±1.7} & 55.3±4.5 & \multicolumn{1}{c|}{53.0±5.5} & \multicolumn{1}{c|}{51.9±4.9} & 48.7±7.8 & \multicolumn{1}{c|}{74.9±1.3} & 74.6±1.3 \\
CB                       & \multicolumn{1}{c|}{71.8±3.4} & \multicolumn{1}{c|}{73.1±3.1} & 71.8±2.4 & \multicolumn{1}{c|}{61.8±3.9} & \multicolumn{1}{c|}{58.3±3.6} & 56.5±5.6 & \multicolumn{1}{c|}{74.6±0.3} & 74.6±0.6 \\
GS                       & \multicolumn{1}{c|}{74.5±1.4} & \multicolumn{1}{c|}{75.1±1.5} & 73.1±1.2 & \multicolumn{1}{c|}{60.3±2.4} & \multicolumn{1}{c|}{55.6±2.1} & 53.0±2.6 & \multicolumn{1}{c|}{73.9±1.1} & 71.8±0.8 \\
RN                       & \multicolumn{1}{c|}{72.2±1.9} & \multicolumn{1}{c|}{73.7±2.3} & 71.4±2.4 & \multicolumn{1}{c|}{57.6±2.5} & \multicolumn{1}{c|}{53.1±2.4} & 53.1±2.4 & \multicolumn{1}{c|}{73.6±0.2} & 70.2±0.7 \\
GM                       & \multicolumn{1}{c|}{74.5±0.4} & \multicolumn{1}{c|}{74.6±0.4} & 74.5±0.2 & \multicolumn{1}{c|}{62.4±0.5} & \multicolumn{1}{c|}{57.2±0.5} & 58.4±0.5 & \multicolumn{1}{c|}{76.3±0.5} & 74.2±0.4 \\ 
GE                       & \multicolumn{1}{c|}{77.8±0.0} & \multicolumn{1}{c|}{72.9±0.2} & 73.1±0.1 & \multicolumn{1}{c|}{66.9±0.2} & \multicolumn{1}{c|}{60.2±0.2} & 58.7±0.3 & \multicolumn{1}{c|}{78.1±0.1} & 74.1±0.4 \\ \midrule
\text{VG}                  & \multicolumn{1}{c|}{\underline{77.9±0.8}} & \multicolumn{1}{c|}{\underline{78.4±0.8}} & \underline{76.6±0.9} & \multicolumn{1}{c|}{\underline{67.1±1.1}} & \multicolumn{1}{c|}{\underline{62.1±0.9}} & \underline{61.6±1.0} & \multicolumn{1}{c|}{\underline{78.2±0.7}} & \underline{77.4±0.9} \\

$\text{VG}^\spadesuit$                  & \multicolumn{1}{c|}{\textit{78.5±1.0}} & \multicolumn{1}{c|}{\textit{79.1±0.9}} & \textit{77.7±1.1} & \multicolumn{1}{c|}{\textit{67.7±1.8}} & \multicolumn{1}{c|}{\textit{62.5±1.4}} & \textit{61.7±1.7} & \multicolumn{1}{c|}{\textit{78.3±0.9}} & \textit{77.6±0.9} \\ \bottomrule
\end{tabular}
\begin{tablenotes}
        \footnotesize
        \item[] \scriptsize Due to space constraints, methods are abbreviated, "VG" is the abbreviation for \model.
    \end{tablenotes}
\end{threeparttable}
\label{exp:main_res_1}
\end{table}
 
\begin{table}[htb]
\caption{The experiment results on PubMed, AmazonPhoto and Wiki-CS.}
\begin{threeparttable}
\begin{tabular}{@{}cc|ccc|ccc@{}}
\toprule
\multirow{2}{*}{Methods}     & PubMed   & \multicolumn{3}{c|}{A-Photo}                                             & \multicolumn{3}{c}{Wiki-CS}                                                                                                 \\ \cmidrule(l){2-8} 
                             & F1       & \multicolumn{1}{c|}{Acc}      & \multicolumn{1}{c|}{bAcc}     & F1       & \multicolumn{1}{c|}{Acc}                       & \multicolumn{1}{c|}{bAcc}                      & F1                        \\ \midrule
\multicolumn{1}{c|}{CE}      & 74.4±0.5 & \multicolumn{1}{c|}{87.6±1.6} & \multicolumn{1}{c|}{81.3±3.7} & 82.9±3.5 & \multicolumn{1}{c|}{70.9±0.2}                  & \multicolumn{1}{c|}{63.1±0.2}                  & 67.9±0.1                  \\
\multicolumn{1}{c|}{RW}      & 74.3±0.9 & \multicolumn{1}{c|}{88.3±5.1} & \multicolumn{1}{c|}{88.7±2.2} & 87.3±4.0 & \multicolumn{1}{c|}{66.5±0.5}                  & \multicolumn{1}{c|}{65.3±0.5}                  & 65.7±0.5                  \\
\multicolumn{1}{c|}{FC}      & 74.7±1.2 & \multicolumn{1}{c|}{85.5±0.3} & \multicolumn{1}{c|}{80.9±0.5} & 81.3±0.4 & \multicolumn{1}{c|}{70.7±0.1}                  & \multicolumn{1}{c|}{63.7±0.3}                  & 68.4±0.3                  \\
\multicolumn{1}{c|}{CB}      & 74.3±0.4 & \multicolumn{1}{c|}{90.4±0.4} & \multicolumn{1}{c|}{88.0±0.5} & 88.1±0.5 & \multicolumn{1}{c|}{71.4±0.4}                  & \multicolumn{1}{c|}{64.0±0.5}                  & 68.4±0.7                  \\
\multicolumn{1}{c|}{GS}      & 73.0±0.8 & \multicolumn{1}{c|}{91.8±0.8} & \multicolumn{1}{c|}{\underline{90.2±1.2}} & \underline{90.2±1.4} & \multicolumn{1}{c|}{67.2±0.4}                  & \multicolumn{1}{c|}{65.4±0.3}                  & 66.9±0.6                  \\
\multicolumn{1}{c|}{RN}      & 72.2±0.5 & \multicolumn{1}{c|}{88.6±2.4} & \multicolumn{1}{c|}{84.8±3.2} & 85.4±3.3 & \multicolumn{1}{c|}{70.6±0.2}                  & \multicolumn{1}{c|}{62.7±0.3}                  & 67.3±0.3                  \\
\multicolumn{1}{c|}{GM}      & 73.1±0.7 & \multicolumn{1}{c|}{89.2±1.1} & \multicolumn{1}{c|}{88.8±1.8} & 86.3±2.0 & \multicolumn{1}{c|}{78.0±0.4} & \multicolumn{1}{c|}{74.7±0.2} & 73.6±0.5 \\
\multicolumn{1}{c|}{GE}      & 74.6±0.1 & \multicolumn{1}{c|}{-}         & \multicolumn{1}{c|}{-}         &     -     & \multicolumn{1}{c|}{-}         & \multicolumn{1}{c|}{-}         &    -      \\ \midrule
\multicolumn{1}{c|}{VG} & \underline{77.3±0.7} & \multicolumn{1}{c|}{\underline{92.0±1.1}} & \multicolumn{1}{c|}{89.6±2.1} & 90.2±1.9 & \multicolumn{1}{c|}{\multirow{2}{*}{\underline{79.4±0.4}}} & \multicolumn{1}{c|}{\multirow{2}{*}{\underline{76.2±0.7}}} & \multirow{2}{*}{\underline{77.0±0.4}} \\
\multicolumn{1}{c|}{$\text{VG}^\spadesuit$ } & \textit{77.5±0.9} & \multicolumn{1}{c|}{\textit{92.3±0.8}} & \multicolumn{1}{c|}{\textit{89.7±1.6}} & \textit{90.6±1.4} & \multicolumn{1}{c|}{}                          & \multicolumn{1}{c|}{}                          &                           \\ \bottomrule
\end{tabular}
\begin{tablenotes}
        \footnotesize
        \item[] \scriptsize "-" indicates that the results not reported in the original paper or conducted in different imbalance setting from this paper. Wiki-CS is originally a imbalance graph, thus the results for ViGraph and $\text{VIGraph}^\spadesuit$ are identical.
    \end{tablenotes}
\end{threeparttable}
\label{exp:main_res_2}
\end{table}

We can derive three observations from the experiment results. \textbf{First}, VIGraph exhibits strong performance across all datasets, surpassing other baselines by a large margin except on Amazon-photo. \model achieved an average improvement of 2.34\% over the second-best results across the five datasets in terms of F1 score. Second, VIGraph demonstrates effectiveness across different backbone models, showcasing its universality and stability. Thirdly, $\text{VIGraph}^\spadesuit$ achieved even higher performance than \model. This observation not only proves that reserved graph characteristics, i.e., the nodes and edges that should be removed, significantly contribute to enhancing the model's learning capacity but also validates that the existing SMOTE-based methods might have limitations in terms of graph construction. 

\subsection{Results Under Different Imbalance Ratios (RQ2)}
\begin{figure}[htb]
  \begin{minipage}[t]{0.32\linewidth}
    \centering
    \includegraphics[width=\textwidth]{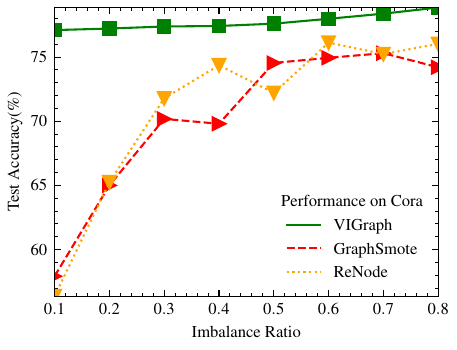}
  \end{minipage}
  \begin{minipage}[t]{0.32\linewidth}
    \centering
    \includegraphics[width=\textwidth]{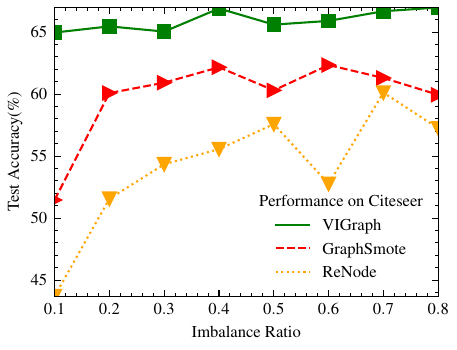}
  \end{minipage}
  \begin{minipage}[t]{0.32\linewidth}
    \centering
    \includegraphics[width=\textwidth]{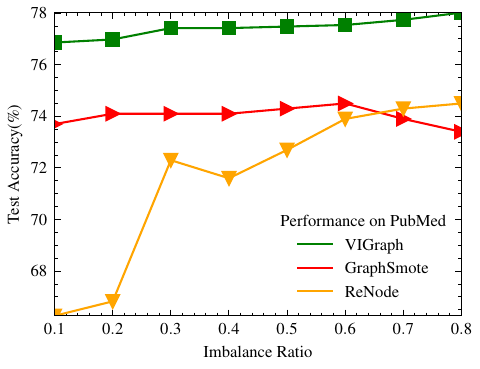}
  \end{minipage}
  \caption{The performance across various imbalance rates.}
  \label{fig:ratio}
\end{figure}
We conducted experiments on Cora, CiteSeer, and PubMed under different imbalance ratios to address \textbf{RQ2}. GraphSmote and ReNode represent the re-sampling and re-weighting methods, respectively, and we chose them for comparison. To investigate the impact of different imbalance ratios ($\lambda$), we varied $\lambda$ from 0.1 to 0.8 with a step size of 0.1. A $\lambda$ value of 0.1 indicates an extremely imbalanced graph, while a value of 0.8 indicates a graph close to a balanced setting. Figure \ref{fig:ratio} presents the results under different imbalance ratios, from which we can observe two key findings. First, VIGraph consistently outperformed GraphSmote and ReNode across various imbalance ratios, demonstrating the universal superiority of VIGraph. Second, VIGraph is more stable than other models under various imbalance ratios, even in extreme imbalance scenarios. This result highlights the robustness of VIGraph and showcases its ability to generate high-quality samples that compensate for the lack of minority nodes.

\subsection{Ablation Studies (RQ3)}
We explore the influence of different training strategies, namely the structure reconstruction strategy, the alignment strategy, and the cross-view contrastive learning strategy, to answer \textbf{RQ3}. Table~\ref{exp:abla} presents the results of the ablation studies conducted on Cora, CiteSeer, and PubMed. The results validate the positive contribution of all three training strategies to the model's performance. However, it is important to note that the impact of each strategy varies across the different datasets, suggesting that each strategy plays a distinct role for different datasets. This highlights the importance of considering dataset-specific characteristics when designing training strategies.
\begin{table*}[htb]
  \centering
  \caption{Ablation Studies of three training strategies.}
  \begin{tabular}{@{}cccccccccc@{}}
    \toprule
    Dataset                   & \multicolumn{3}{c}{Cora} & \multicolumn{3}{c}{CiteSeer} & \multicolumn{3}{c}{PubMed}                                                                                                       \\ \midrule
    Metric                    & Acc                      & bACC                         & F1                         & Acc            & bACC           & F1             & Acc            & bACC           & F1             \\ \midrule
    \model          & \textbf{77.9}           & \textbf{78.4}               & \textbf{76.6}             & \textbf{67.1} & \textbf{62.1} & \textbf{61.6} & \textbf{78.2} & \textbf{77.4} & \textbf{77.3} \\
    w.o. $\mathcal{L}_{rec}$  & 76.3                    & 77.4                        & 75.4                      & 63.8          & 63.1          & 61.3          & 75.0          & 75.2          & 74.4          \\
    w.o. $\mathcal{L}_{elbo}$ & 77.0                    & 78.5                       & 76.8                      & 65.1          & 60.9          & 60.5          & 76.6          & 76.8          & 76.1          \\
    w.o. $\mathcal{L}_{gcl}$  & 75.5                    & 76.8                        & 75.0                      & 63.0          & 59.8          & 59.2           & 75.3          & 75.6          & 74.7          \\ \bottomrule
  \end{tabular}
  \label{exp:abla}
\end{table*}

\section{CONCLUSION AND DISCUSSION}
In this paper, we focus on addressing class-imbalanced node classification using a generative model. We analyze the limitations of existing methods, particularly their flawed imbalanced graph construction. Then, we propose VIGraph as a generative SSL method to tackle this problem. VIGraph develops comprehensive strategies to facilitate the overall training process. Furthermore, \model utilizes variational inference to directly generate nodes for minority classes.
Extensive experiments demonstrate the effectiveness and robustness of VIGraph. 
\bibliographystyle{splncs04}
\bibliography{pkdd}

\end{document}

%% file: math_commands.tex
\usepackage{amsmath,amsfonts,bm}

\def\eqref#1{equation~\ref{#1}}

\def\1{\bm{1}}

\DeclareMathAlphabet{\mathsfit}{\encodingdefault}{\sfdefault}{m}{sl}
\SetMathAlphabet{\mathsfit}{bold}{\encodingdefault}{\sfdefault}{bx}{n}

